\pdfoutput=1

\documentclass[11pt]{article}

\usepackage[]{ACL2023}

\usepackage{times}
\usepackage{latexsym}
\usepackage{algpseudocode}
\usepackage{algorithm}
\usepackage{graphicx}
\usepackage{hyperref}
\usepackage{amsmath} 
\usepackage{spverbatim} 
\usepackage{amsfonts}
\usepackage{booktabs}

\usepackage[T1]{fontenc}

\usepackage[utf8]{inputenc}

\usepackage{microtype}

\usepackage{inconsolata}

\title{GateNLP at SemEval-2025 Task 10: Hierarchical Three-Step Prompting for Multilingual Narrative Classification}

\author{Iknoor Singh, Carolina Scarton {\normalfont and} Kalina Bontcheva \\
  Department of Computer Science, University of Sheffield (UK)\\
  \texttt{\{i.singh, c.scarton, k.bontcheva\}@sheffield.ac.uk} \\}

\begin{document}
\maketitle
\begin{abstract}
The proliferation of online news and the increasing spread of misinformation necessitate robust methods for automatic data analysis. Narrative classification is emerging as a important task, since identifying what is being said online is critical for fact-checkers, policy markers and other professionals working on information studies. 
This paper presents our approach to SemEval 2025 Task 10 Subtask 2, which aims to classify news articles into a predefined two-level taxonomy of main narratives and sub-narratives across multiple languages. We propose  Hierarchical Three-Step Prompting (\texttt{H3Prompt}) for multilingual narrative classification. Our methodology follows a three-step Large Language Model (LLM) prompting strategy, where the model first categorises an article into one of two domains (Ukraine-Russia War or Climate Change), then identifies the most relevant main narratives, and finally assigns sub-narratives. Our approach secured the top position on the English test set among 28 competing teams worldwide. 
The code is available at \url{https://github.com/GateNLP/H3Prompt}.

\end{abstract}

\section{Introduction}

The rapid dissemination of information online has significantly influenced public discourse, making it crucial to detect and classify narratives accurately \cite{heinrich2024automatic, piskorski2022exploring}. Narrative classification plays a key role in understanding how different perspectives shape public opinion and in identifying potential misinformation campaigns \cite{amanatullah2023tell}. To advance research in this area, the SemEval 2025 shared task 10 \cite{semeval2025task10} presents multilingual characterisation and extraction of narratives from online news providers. A narrative is defined as a structured presentation of information that conveys a specific message or viewpoint, often forming a cohesive storyline\footnote{\url{https://www.merriam-webster.com/dictionary/narrative}}. The task provides a benchmark for evaluating and developing narrative classification models \cite{semeval2025task10}, helping researchers analyse how narratives emerge and propagate across different languages.

As part of this challenge, Subtask 2 \cite{semeval2025task10} focuses on assigning appropriate sub-narrative labels to a given news article based on a two-level taxonomy\footnote{\url{https://propaganda.math.unipd.it/semeval2025task10/NARRATIVE-TAXONOMIES.pdf}} \cite{guidelinesSE25T10}, where each narrative is further divided into sub-narratives. This is a multi-label, multi-class document classification task involving news articles from two key domains: the Ukraine-Russia war and climate change. The dataset comprises articles, collected between 2022 and mid-2024, in five languages: Bulgarian, English, Hindi, Portuguese, and Russian. A significant portion of these articles have been flagged by fact-checkers as potentially spreading misinformation \cite{semeval2025task10}.

Previous work has focused on fine-grained narrative classification across various domains, including climate change \cite{coan2021computer, piskorski2022exploring, zhou2024large, rowlands2024predicting}, the Ukraine-Russia war \cite{amanatullah2023tell}, health misinformation \cite{ganti2023narrative}, and the COVID-19 infodemic \cite{kotseva2023trend, heinrich2024automatic, shahsavari2020conspiracy}. These studies have proposed models to identify narratives, aiding in the analysis of misinformation and public discourse within these critical topics.

Given the complexity and multilingual nature of this task, this paper proposes a novel Hierarchical Three-Step Prompting (\texttt{H3Prompt}). In this, we fine-tune a Large Language Model (LLM) from the \texttt{LLaMA 3.2} family using \texttt{H3Prompt} by leveraging both training data and synthetically generated data. Our method follows a three-step prompting framework, ensuring a structured and hierarchical classification process. This approach enhances the model’s ability to accurately distinguish between narratives and sub-narratives, improving classification performance across multiple languages. Moreover, it also allows analysts to gain deeper insights into emerging narratives.

\section{Hierarchical Three-Step Prompting (\texttt{H3Prompt})}
Our approach to narrative classification follows a hierarchical three-step prompting mechanism. We first describe the dataset and synthetic data generation process (Section \ref{dataset}). Next, we outline fine-tuning details (Section \ref{lora}). Finally, we detail the prompt structure for refining predictions across classification levels (Section \ref{H3Prompt}).

\subsection{Dataset}
\label{dataset}
We utilise the training dataset provided by SemEval 2025 task organisers, which includes annotated news articles spanning five languages \cite{semeval2025task10}. We translate all non-English articles into English using Fairseq's \texttt{m2m100\_418M} model \citep{fan2021beyond}. After translation, we obtain a total of 2,091 annotated data points. 

Additionally, we synthetically generate articles to augment the dataset in order to improve model generalisation. We used an Vicuna LLM \cite{zheng2023judging} to generate synthetic articles.

\noindent{\textbf{Used prompt:}}
\begin{spverbatim}
You are an AI news curator. Generate 5 different news articles related to the following topic on {category}.

Topic: {sub_narrative}
Explanation: {explanation}

Each article should be between 400-500 words and explore a unique aspect, perspective, or event related to this topic. Focus on delivering informative, coherent, and engaging articles that reflect diverse points of view or angles on the given topic. Avoid redundancy by ensuring that each article highlights a different aspect or argument related to the context provided. The output format should look like this:
Article 1:
Article 2:
Article 3:
Article 4:
Article 5:
\end{spverbatim}

We opt for \texttt{vicuna-7b-v1.5} \cite{zheng2023judging} for synthetic data generation since it has been shown to easily generate content containing disinformation \cite{vykopal2023disinformation}. As shown in the prompt, we provide both the narrative and its explanation to the model. We generate explanations using ChatGPT and manually verify them (see Appendix \ref{appendix}).

We generate 100 articles for each sub-narrative. To encourage diversity, we generate articles using sampling with different temperature values in the range of 1 to 1.5. In total, we synthetically generate 8,129 news articles.

Finally, a total of 10,220 (2,091 + 8,129) news articles, including both annotated and synthetic data, are used for training the models.

\subsection{Low-Rank Adaptation Fine-Tuning}
\label{lora}
Low-Rank Adaptation (LoRA) was introduced by \citet{hu2021lora} and applied specifically to the attention layers of transformer models. This approach demonstrated comparable or superior performance to full fine-tuning while significantly reducing the number of trainable parameters.

The pre-trained transformer consists of multiple dense layers, where the transformation of an input vector \( x \) into an output representation \( h \) is performed through full-rank matrix multiplication. In a standard pre-trained model, this transformation is represented as follows:
\[
h = W_0 x
\]
where \( W_0 \in \mathbb{R}^{d \times k} \) is the original pre-trained weight matrix. 
During model adaptation in LoRA, fine-tuning introduces weight modifications, allowing the updated output to be expressed as:
\[
h_{\text{adapted}} = W_0 x + \Delta W x
\]
where \( \Delta W \) represents the learned weight adjustments optimised through training.
LoRA constrains these weight updates by decomposing \( \Delta W \) into two lower-rank matrices: \( B \in \mathbb{R}^{d \times r} \) and \( A \in \mathbb{R}^{r \times k} \), where \( r \ll \min(d, k) \). This formulation allows the adapted output to be computed as:
\[
h_{\text{LoRA}} = W_0 x + B A x
\]
Matrices \( A \) and \( B \) are the trainable parameters, initialised such that their product \( B A \) starts as a zero matrix. During training, original pre-trained weight matrix \( W_0 \) is frozen and does not receive gradient updates. Additionally, the weight update \( \Delta W x \) is scaled by a factor of \( \frac{\alpha}{r} \), where \( \alpha \) is a hyperparameter controlling the adaptation strength.  

In this paper, we use LoRA to fine-tune \texttt{LLaMA-3.2-3B-Instruct}  \cite{dubey2024llama, touvron2023llama} using the Unsloth library \cite{unsloth}. The fine-tuning process is guided by the prompts defined in Section \ref{H3Prompt}. We set \( \alpha \) and \( r \) to 64,  the number of epochs to 5, the batch size to 8, the gradient accumulation steps to 8, and the learning rate to $2e-4$. We manually tune the hyperparameters within the following bounds: (i) 1 to 8 epoch (ii) $1e-5$ to $5e-4$ learning rate (iii) 2 to 16 batch size (iv) 8 to 128 for both \( \alpha \) and \( r \) values. All experiments are conducted on three NVIDIA A100 40GB GPUs.

\subsection{Prompting Mechanism for Narrative Classification}
\label{H3Prompt}

In this subsection, we elaborate on the detailed structure of the \texttt{H3Prompt} mechanism. This includes the prompts employed at each step and the accompanying algorithm to do the classification.

\paragraph{Step 1: Category Classification.} The first step determines whether a document belongs to the ``Ukraine-Russia War'' or ``Climate Change'' category. If no match is found, the document is assigned the label ``Other.'' This first step filters out all irrelevant news articles.

\noindent{\textbf{Used prompt:}}
\begin{spverbatim}
Given the following document text, classify it into one of the two categories: "Ukraine-Russia War" or "Climate Change".

Document Text: {document_text}

Determine the category that closely or partially fits the document. If neither category applies, return "Other". Return only the output, without any additional explanations or text.
\end{spverbatim}

\paragraph{Step 2: Main Narrative Classification.} Based on the assigned category in Step 1, \texttt{H3Prompt} then selects the most relevant main narratives using a predefined taxonomy with explanations for each main narrative. See Appendix \ref{appendix} for explanation details. 
The model returns one or more main narratives as hash-separated labels. If no relevant narrative is found, "Other" is returned.

\noindent{\textbf{Used prompt:}}
\begin{spverbatim}
The document text given below is related to "{category}".
Please classify the document text into the most relevant narratives. Below is a list of narratives along with their explanations:

{narratives_list_with_explanations}

Document Text: {document_text}

Return the most relevant narratives as a hash-separated string (e.g., Narrative1#Narrative2..). If no specific narrative can be assigned, just return "Other" and nothing else. Return only the output, without any additional explanations or text.
\end{spverbatim}





\paragraph{Step 3: Sub-Narrative Classification.} For each identified main narrative (Step 2), \texttt{H3Prompt} assigns relevant sub-narratives by leveraging a structured prompt that includes explanations of available sub-narratives. See Appendix \ref{appendix} for details on how these explanations were generated. Only the sub-narratives corresponding to the main narratives identified in Step 2 are used in the prompt.
If no suitable sub-narrative is found, "Other" is returned.

\noindent{\textbf{Used prompt:}}
\begin{spverbatim}
The document text given below is related to "{category}" and its main narrative is: "{main_narrative}".
Please classify the document text into the most relevant sub-narratives. Below is a list of sub-narratives along with their explanations:

{sub_narratives_list_with_explanations}

Document Text: {document_text}

Return the most relevant sub-narratives as a hash-separated string (e.g., Sub-narrative1#Sub-narrative2..). If no specific sub-narrative can be assigned, just return "Other" and nothing else. Return only the output, without any additional explanations or text.
\end{spverbatim}

The systematic pseudocode for classifying news articles into narratives and sub-narratives is presented in Algorithm \ref{alg:narrative_classification}.

\begin{algorithm}
\caption{Hierarchical Three-Step Prompting}
\label{alg:narrative_classification}
\begin{algorithmic}[1]
\Require Document text $D$
\Require Narrative taxonomy $T$ with main narratives $N_m$ and sub-narratives $N_s$
\Ensure Assigned category, main narratives, and sub-narratives

\State $category \gets \textsc{ClassifyCategory}(D)$
\If{$category == \text{Other}$}
\State \Return ${\text{Other}}$
\EndIf

\State $mainNarratives \gets \textsc{MainNarrative}(D)$
\If{$mainNarratives == \text{Other}$}
\State \Return ${\text{Other}}$
\EndIf

\State $labels \gets \emptyset$
\For{each $n_m \in mainNarratives$}
\State $subNarratives \gets \textsc{SubNarrative}(D)$
\For{each $n_s \in subNarratives$}
\If{$n_s \in N_s$}
\State $labels \gets labels \cup {(n_m, n_s)}$
\Else
\State $labels \gets labels \cup {(n_m, \text{Other})}$
\EndIf
\EndFor
\EndFor

\Return $labels$
\end{algorithmic}
\end{algorithm}

\begin{table*}[]
\centering

\resizebox{\textwidth}{!}{%
\begin{tabular}{@{}lcccc@{}}
\toprule
\textbf{Method}                                 & \multicolumn{1}{l}{\textbf{F1 Macro Coarse}} & \multicolumn{1}{l}{\textbf{F1 Macro Coarse (STD)}} & \multicolumn{1}{l}{\textbf{F1 Samples Fine}} & \multicolumn{1}{l}{\textbf{F1 Samples Fine (STD)}} \\ \midrule
\textbf{Zero-shot Models}                                    &                                         &                                               &                                         &                                              \\ \midrule
GPT-4o-mini                                     & 0.456                                        & 0.343                                              & 0.291                                        & 0.278                                              \\
GPT-4o                                          & 0.465                                        & 0.374                                              & 0.286                                        & 0.304                                              \\
LLaMA-3.2-3B-Instruct                           & 0.249                                        & 0.313                                              & 0.167                                        & 0.275                                              \\
LLaMA-3.1-8B-Instruct                           & 0.237                                        & 0.332                                              & 0.159                                        & 0.276                                              \\
FuseChat-LLaMA-3.2-3B-Instruct                  & 0.225                                        & 0.319                                              & 0.160                                        & 0.283                                              \\
Gemma-2-2b-it                                   & 0.324                                        & 0.413                                              & 0.278                                        & 0.402                                              \\
Random Baseline                                 & 0.106                                        & 0.267                                              & 0.000                                        & 0.000                                              \\ \midrule
\textbf{Trained Models }                                    &                                         &                                               &                                         &                                              \\ \midrule
Logistic Regression                             & 0.260                                        & 0.433                                              & 0.260                                        & 0.433                                              \\
LightGBM                                        & 0.434                                        & 0.434                                              & 0.352                                        & 0.440                                              \\
RoBERTa-base (B)                                     & 0.490                                        & 0.387                                              & 0.383                                        & 0.403                                              \\
RoBERTa-base (w/o synth)                                     & 0.529                                        & 0.375                                              & 0.397                                        & 0.354                                              \\
RoBERTa-base                                         & 0.543                                        & 0.376                                              & 0.439                                        & 0.378                                              \\
LLaMA-3.2-3B-Instruct (B)                       & 0.562                                        & 0.409                                              & 0.428                                        & 0.380                                            
  \\ \midrule
\textbf{H3Prompt models}                                    &                                         &                                               &                                         &                                              \\ \midrule
LLaMA-3.2 H3Prompt (w/o synth)                       & 0.502                                        & 0.394                                              & 0.392                                        & 0.369                                              \\
LLaMA-3.2 H3Prompt                          & 0.577                                        & 0.390                                              & 0.482                                        & 0.390                                              \\
LLaMA-3.2 H3Prompt (Ensemble - Union)        & \textbf{0.623}                               & 0.352                                              & \textbf{0.516}                               & 0.364                                              \\
LLaMA-3.2 H3Prompt (Ensemble - Majority Vote)           & 0.567                                        & 0.410                                              & 0.482                                        & 0.404                                              \\
LLaMA-3.2 H3Prompt (Ensemble - Intersection) & 0.458                                        & 0.432                                              & 0.401                                        & 0.409                                              \\ \bottomrule
\end{tabular}%
}
\caption{F1 score results for coarse- and fine-grained classification on the development set (English only). STD is the standard deviation of samples F1 score. \textbf{w/o synth} indicates that the model is trained only on the provided training data (i.e., without synthetic data), and \textbf{B} denotes that the model is trained using binary classification only. The best results are in bold.}
\label{tab:my-table}
\end{table*}

\section{Experimental Details}

\subsection{Baseline Models}
To assess the performance of our \texttt{H3Prompt}, we test a range of baseline models. We also experiment with different configurations: binary classification (denoted by \textbf{B}), in which training and classification are performed for each sub-narrative separately; and models trained exclusively on the annotated data provided by the shared task organisers without synthetic data (denoted by \textbf{w/o synth}).

\paragraph{Random Baseline.} Provided by the organisers \cite{semeval2025task10}, it randomly assigns labels based on the training dataset's distribution.

\paragraph{Traditional Machine Learning.} We implement logistic regression and LightGBM using TF-IDF features as input embeddings. 

\paragraph{Zero-shot Models.} We evaluate several LLMs such as \texttt{GPT-4o}, \texttt{GPT-4o-mini}, \texttt{LLaMA-3.2-3B-Instruct}, \texttt{LLaMA-3.1-8B-Inst\\ruct}, \texttt{FuseChat-Llama-3.2-3B-Instruct}, and \texttt{Gemma-2-2B-it} in a zero-shot setting. We use the same prompts as those described in Section \ref{H3Prompt}.

\paragraph{Fine-tuned Transformer Models.} We train \texttt{RoBERTa-base} models using different configurations, including binary classification and with or without synthetic data. 
For \texttt{RoBERTa-base}, a three-step classifier is used to predict the category, main narrative, and sub-narrative. We set the learning rate to $1e-5$, the batch size to 32, and epochs to 4. The label is selected based on an output threshold, which is manually tuned in the range of 0.2 to 0.8.


\section{Results and Discussion}

Table \ref{tab:my-table} presents the F1 scores for various baseline and fine-tuned models on the development set of English. The official evaluation measure for the task is samples F1 score for sub-narratives (fine-grained) and macro F1 for narratives (coarse-grained) \cite{semeval2025task10}.

In zero-shot models, \texttt{GPT-4o} achieves the highest F1-score for coarse-grained classification and \texttt{GPT-4o-mini} gives the highest F1-score for fine-grained classification. On the other hand,
zero-shot models, such as \texttt{LLaMA-3.2-3B-Instruct} and \texttt{Gemma-2-2b-it}, perform significantly worse than trained models, indicating that domain-specific fine-tuning is crucial for improving the narrative classification performance. 

Among trained models, logistic regression and LightGBM achieve moderate performance, but transformer-based models such as \texttt{RoBERTa-base} and \texttt{LLaMA-3.2 H3Prompt} outperforms them. Notably, our hierarchical three-step prompting approach (\texttt{LLaMA-3.2 H3Prompt}) achieves an F1 Macro Coarse score of 0.577 and an F1 Samples Fine score of 0.482, demonstrating the effectiveness of structured classification. 

The results also indicate that incorporating synthetic data during training improves performance, as models trained solely on the provided training data (denoted by \textbf{w/o synth}) perform worse than those that incorporate additional synthetic data. For instance, for \texttt{LLaMA-3.2 H3Prompt}, training with synthetic data improves the fine-grained F1 score by 23\% (improvement from 0.392 to 0.482), while for \texttt{RoBERTa-base}, it leads to a 10\% improvement (improvement from 0.397 to 0.439).

In addition, binary classification models (denoted by \textbf{B}) showed a slight decrease in performance compared to hierarchical prompting models, reinforcing the importance of a structured three-step classification approach. 

To further improve classification, we use the best-performing model (i.e., \texttt{LLaMA-3.2 H3Prompt}) to experiment with a bagging ensemble \cite{breiman1996bagging} to reduce variance from individual models. Specifically, we train three different models on separate subsets of the dataset and then combine their predictions. We use three different strategies to aggregate the predictions: (1) union-based, where a sub-narrative is selected if any model predicts it; (2) majority-vote, where a sub-narrative is selected if at least two of the models predict it; and (3) intersection-based, where a sub-narrative is selected only if all models predict it.

Among the ensemble methods, we find that \texttt{LLaMA-3.2 H3Prompt (Ensemble - Union)} is the best-performing model. It achieved the highest scores, with 0.623 for narratives and 0.516 for sub-narratives, showcasing the advantage of ensemble methods in improving classification robustness.

Furthermore, we submitted our best-performing run, \texttt{LLaMA-3.2 H3Prompt (Ensemble - Union)}, for evaluation on the test set. We submitted our test predictions for Bulgarian, English, Hindi, Portuguese, and Russian. For all non-English articles, we first machine-translated\footnote{We use \texttt{m2m100\_418M} \citep{fan2021beyond} for translation.} them into English and then used the translated text for inference.
As shown on the test leaderboard\footnote{\url{https://propaganda.math.unipd.it/semeval2025task10/leaderboardv3.html}}, our \texttt{GATENLP} submission secured 1st place for English, Portuguese, and Russian. For Bulgarian and Hindi, it ranked 3rd and 5th, respectively. These results highlight the potential of our method for fine-grained narrative classification across multiple languages and misinformation domains.

\section{Conclusion}
In this paper, we introduced Hierarchical Three-Step Prompting (\texttt{H3Prompt}) for multilingual narrative classification as part of SemEval 2025 Task 10 Subtask 2. Our approach fine-tuned \texttt{LLaMA 3.2} using both annotated training data and synthetically generated news articles to enhance classification robustness. Our method secured the top position on the English test set among 28 competing teams worldwide, demonstrating the effectiveness of our approach for fine-grained narrative classification. Experimental results showed that \texttt{H3Prompt} outperforms baseline methods and zero-shot models, achieving state-of-the-art performance in narrative and sub-narrative classification. We further demonstrated that incorporating synthetic data during training significantly improves model performance. Additionally, ensemble methods provided further enhancements, achieving the highest scores across multiple languages.


\section*{Acknowledgements}
This work is supported by the UK’s innovation agency (InnovateUK) grant number 10039039 (approved under the Horizon Europe Programme as VIGILANT, EU grant agreement number 101073921) (https://www.vigilantproject.eu).

We would also like to thank Ibrahim Abu Farha and Fatima Haouari for the useful discussions regarding the task.

\bibliography{anthology,custom}
\bibliographystyle{acl_natbib}

\appendix

\section{Narrative Explanation}
\label{appendix}

To generate explanations for the main narratives and sub-narratives, we used ChatGPT. Specifically, we prompted the model to generate explanations:

\noindent{\textbf{Used prompt:}}
\begin{spverbatim}
You are given main narratives and sub-narratives for the Ukraine-Russia War and Climate Change. Now, provide a concise explanation for each main narrative and its sub-narratives.

{main_narratives}
{sub_narratives}

\end{spverbatim}

The generated explanations were manually reviewed and refined to ensure clarity and accuracy. The final set of narrative explanations used in our classification experiments is available at:
\url{https://github.com/GateNLP/H3Prompt/tree/master/Dataset}.

\end{document}